\documentclass[lettersize,journal]{IEEEtran}
\IEEEoverridecommandlockouts
\usepackage{amsmath,amssymb,amsfonts}
\usepackage{algorithm}
\usepackage{algpseudocode} 
\usepackage{array}
\usepackage[caption=false,font=normalsize,labelfont=sf,textfont=sf]{subfig}
\usepackage{textcomp}
\usepackage{xcolor}
\def\BibTeX{{\rm B\kern-.05em{\sc i\kern-.025em b}\kern-.08em
    T\kern-.1667em\lower.7ex\hbox{E}\kern-.125emX}}
\usepackage{stfloats}
\usepackage{url}
\usepackage{verbatim}
\usepackage{graphicx}
\usepackage{cite}
\usepackage{booktabs}
\usepackage{bbding}
\usepackage{authblk}
\usepackage[backref]{hyperref} 
\hypersetup{
hidelinks
}
\usepackage{array}
\newcommand{\PreserveBackslash}[1]{\let\temp=\\#1\let\\=\temp}
\newcolumntype{C}[1]{>{\PreserveBackslash\centering}p{#1}}
\begin{document}

\title{Heterogeneous Feature Distillation Network for SAR Image Semantic Segmentation }

\begin{comment}
\author{Mengyu Gao}
\author{Quilei Dong\textsuperscript{1}}

\thanks{The authors are with the National Laboratory of Pattern Recognition, Institute of Automation, Chinese Academy of Sciences, Beijing 100190, China, also with the School of Artificial Intelligence, University of Chinese Academy of Sciences, Beijing 100049, China,}
\end{comment}

\author{Mengyu Gao and Qiulei Dong
        % <-this % stops a space
\thanks{Corresponding author: Qiulei Dong}
\thanks{Mengyu Gao and Qiulei Dong are with the National Laboratory of Pattern Recognition, Institute of Automation, Chinese Academy of Sciences, Beijing 100190, China, also with the School of Artificial Intelligence, University of Chinese Academy of Sciences, Beijing 100049, China, and also with the Center for Excellence in Brain Science and Intelligence Technology, Chinese Academy of Sciences, Beijing 100190, China (e-mail: gaomengyu2021@ia.ac.cn; qldong@nlpr.ia.ac.cn).}}



\maketitle

\begin{abstract}
Semantic segmentation for SAR (Synthetic Aperture Radar) images has attracted increasing attention in the remote sensing community recently, due to SAR's all-time and all-weather imaging capability. However, SAR images are generally more difficult to be segmented than their EO (Electro-Optical) counterparts, since speckle noises and layovers are inevitably involved in SAR images. To address this problem, we investigate how to introduce EO features to assist the training of a SAR-segmentation model, and propose a heterogeneous feature distillation network for segmenting SAR images, called HFD-Net, where a SAR-segmentation student model gains knowledge from a pre-trained EO-segmentation teacher model. In the proposed HFD-Net, both the student and teacher models employ an identical architecture but different parameter configurations, and a heterogeneous feature distillation model is explored for transferring latent EO features from the teacher model to the student model and then enhancing the ability  of the student model for SAR image segmentation. In addition, a heterogeneous feature alignment module is explored to aggregate multi-scale features for segmentation in each of the student model and teacher model. Extensive experimental results on two public datasets demonstrate that the proposed HFD-Net outperforms seven state-of-the-art SAR image semantic segmentation methods.
\end{abstract}


\section{Introduction}
\label{1}
\IEEEPARstart{S}{AR} (Synthetic Aperture Radar) image semantic segmentation, which aims at automatically assigning a label to each pixel, is a challenging topic in the remote sensing community. In recent years, it has played an important role in various tasks, such as oil spill detection\cite{oilspilldetection3}, glacial melting detection\cite{glacierdetection4}, geological hazard assessment\cite{geological3}, etc.

\par According to whether deep neural networks (DNN) are used or not, the existing semantic segmentation methods for SAR images could be divided into two categories: traditional segmentation methods and DNN-based segmentation methods. Traditional SAR segmentation methods usually utilize some traditional machine learning techniques to segment SAR images, such as markov random field \cite{traditionalCRF} and probabilistic model \cite{probabilistic}, while DNN-based SAR segmentation methods generally employ various neural network architectures for handling the SAR segmentation task \cite{finetuneFCN},\cite{glacierdetection2},\cite{finetuneintro1}, \cite{multimodalintro1}, \cite{finetuneintro2}.

\par Recently, due to the rapid development of deep learning, DNN-based SAR segmentation methods have attracted increasing attention. A straightforward strategy is to fine-tune the existing EO (Electro-Optical) segmentation networks with SAR images\cite{finetuneFCN}, \cite{finetuneintro2}, inspired by the DNN's success in segmenting EO images\cite{EOintro1}, \cite{EOintroUNet}. However, such a strategy only has limited effectiveness, due to the fact that SAR and EO images have different imaging mechanisms. To alleviate this problem, some works\cite{SARsegintro3},\cite{SARsegintro1} for SAR semantic segmentation have been proposed by utilizing the intrinsic characteristics in SAR images. But it is still difficult for these methods to achieve competitive segmentation results in comparison to the EO image segmentation results by some state-of-the-art EO segmentation methods\cite{EOintroDeepLabv3}, \cite{EOintroRefineNet} because of two main reasons: (i) SAR images generally contain speckle noises and layovers and (ii) EO images have more abundant textures than SAR images.

\begin{figure}[!t]
\centerline{\includegraphics[scale=0.35]{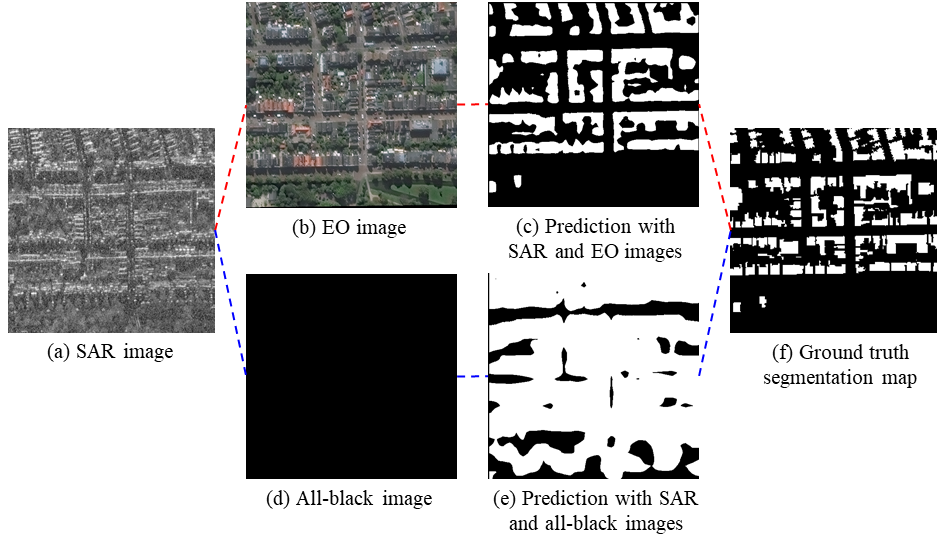}}
\caption{EO images' influence on the performance of the typical multi-modal method MCANet\cite{MCANet} for segmenting buildings, which is trained with 2401 pairs of SAR and EO images from the SpaceNet6 dataset\cite{SpaceNet6}. MCANet is tested by utilizing two kinds of input image pairs respectively, including (i) a pair of SAR image and its corresponding EO image (denoted in red dotted line); (ii) a pair of SAR image and an all-black image which indicates no EO scene image is available (denoted in blue dotted line):  (a) A testing SAR image; (b) The corresponding EO image to (a); (c) Predicted segmentation map with the pair of SAR (a) and EO (b) images; (d) An all-black image; (e) Predicted segmentation map with the pair of SAR (a) and all-black (e) images; (f) Ground truth segmentation map. As is seen, once EO image is not used, the segmentation performance of MCANet degraded significantly.}
\label{Figure 1}
\end{figure}

\par The above issue further encourages researchers to design multi-modal methods for simultaneously segmenting both SAR and EO images\cite{multimodalintro1}, \cite{multimodalintro2}. These multi-modal methods use pairs of SAR and EO images not only for training but also for testing. However, once they segment SAR images singly without EO images, their segmentation performances would be degraded significantly. Fig. \ref{Figure 1} illustrates this issue by evaluating a typical multi-modal method MCANet\cite{MCANet} for segmenting buildings from the public dataset SpaceNet6\cite{SpaceNet6} (more results by the other multi-modal methods\cite{CMC}, \cite{RSDNet} in Fig. \ref{Figure 5} are also consistent with those by MCANet\cite{MCANet} in this figure). As seen from this figure,  MCANet\cite{MCANet} is firstly trained with 2401 pairs of SAR and EO images from the SpaceNet6 dataset\cite{SpaceNet6}. Then, it is not only tested by utilizing a pair of SAR and EO images as inputs, but also tested by utilizing a pair of SAR and all-black images as inputs, indicating that only SAR images are available for testing in this case. The segmentation result by utilizing the input SAR image without its corresponding EO image in Fig. \ref{Figure 1}(e) is significantly worse than that by utilizing a pair of SAR and EO images in Fig. \ref{Figure 1}(c). Here, it has to be further pointed out that it is not technically hard to collect pairs of SAR and EO images for off-line training, but in many real testing scenarios (e.g., night, cloud cover, etc.), it is only available to capture SAR images, but generally impossible to capture SAR images and their corresponding clear EO images simultaneously. This motivates us to investigate the following problem: How to jointly use SAR and EO images to train such a SAR segmentation network that could segment SAR images singly without EO images more effectively? 

\par To address this problem, in this paper, we propose a Heterogeneous Feature Distillation Network for single SAR semantic segmentation, called HFD-Net, which is inspired by the knowledge transfer ability of the knowledge distillation technique in other visual tasks, such as object classification \cite{DistillClassification1}, object detection \cite{DistillDetection2}, etc. The proposed HFD-Net, which aims to learn heterogeneous features for segmenting SAR images, consisting of a pre-trained teacher model for EO image segmentation, a student model for SAR image segmentation, and a designed heterogeneous feature distillation model for knowledge transfer. The student model has an identical architecture as its teacher model but a different parameter configuration, and the heterogeneous feature distillation model is designed for transferring latent EO features from the teacher model to the student model so that the performance of the student model on SAR image segmentation could be boosted. Moreover, it is noted that various multi-scale feature aggregation techniques have demonstrated their effectiveness for learning more discriminative features in some other tasks (such as monocular depth estimation\cite{AlignmentDepthEstimation} and visual localization\cite{AlignmentVisualLocalization}), however, all these feature aggregation techniques could only aggregate homogeneous features, but fail to deal with heterogeneous features. Unlike them, a heterogeneous feature alignment module is further designed in this work to aggregate multi-scale heterogeneous features in the student model, which is also used to aggregate multi-scale homogeneous features in the teacher model. 

\par In sum, our main contributions include:
\begin{itemize}
\item We propose the HFD-Net for SAR image semantic segmentation, which could learn heterogeneous EO/SAR features through knowledge distillation. To our best knowledge, this work is the first attempt to segment SAR images with distilled heterogeneous features in the remote sensing field. 
\item We explore the heterogeneous feature distillation model in the proposed HFD-Net, which could automatically transfer knowledge from the EO-segmentation teacher model to the SAR-segmentation student model.
\item We explore the heterogeneous feature alignment module. Unlike the existing feature aggregation techniques\cite{FaPN}\cite{AlignSeg}, this module could not only aggregate multi-scale homogeneous features, but also multi-scale heterogeneous features. 
\end{itemize}

\par The rest of this paper is organized as follows: Section \ref{2} gives a review of the related work on DNN-based SAR semantic segmentation. Section \ref{3} describes the proposed HFD-Net in detail. Extensive experimental results are reported in Section \ref{4}. We conclude the paper in Section \ref{5}.

\section{Related Work}
\label{2}
\par In this section, we review the single-modal and multi-modal DNN-based methods for SAR image semantic segmentation respectively.
\subsection{Single-modal Methods for SAR Semantic Segmentation}
\par Single-modal methods for SAR semantic segmentation use SAR images singly for both training and testing. Many early DNN-based single-modal methods focused on fine-tuning the existing EO segmentation networks with SAR images\cite{finetuneFCN}, \cite{finetunerelate1}, \cite{finetunerelate2}, \cite{finetunerelate3}, \cite{finetuneSegNet} and \cite{finetuneDeepLabv3+}. Bianchi et al.\cite{finetuneFCN} adopted a Fully Convolutional Network (FCN)\cite{EOintroFCN} to detect the presence of avalanches by segmenting SAR images with snow avalanche. Pham et al. \cite{finetuneSegNet} utilized SegNet\cite{EOintroSegNet} for pixel-wise classification over very high resolution airborne PolSAR images. Tom at al. \cite{finetuneDeepLabv3+} addressed lake ice detection using Sentinel-1 SAR data by DeepLabv3+ \cite{EOintroDeepLabv3+}. Holzmann et al.\cite{AttentionUNet} introduced an attention mechanism into the typical U-Net\cite{EOintroUNet} for segmenting SAR images and Davari et al.\cite{MCC} employed a distance map in U-Net\cite{EOintroUNet} to add contextual information.

\par Unlike the above methods that directly utilized the existing EO segmentation networks, some methods investigated new network architectures by introducing intrinsic characteristics in SAR images\cite{SARsegrelate1}, \cite{graphnet}, \cite{SD}, \cite{SARsegrelate3}, \cite{HRSARNet} and \cite{SARsegrelate2}. Wang et al. \cite{HRSARNet} proposed HR-SAR-Net under pyramid structure with atrous convolution to extract magnitude information and phase information separately in SAR images. Liu et al. \cite{graphnet} developed a dark spot detection method based on super-pixels deeper graph convolutional networks (SGDCN) to smooth SAR image noises. Ristea et al. \cite{SD} applied sub-aperture decomposition (SD) algorithm as a preprocessing stage for an unsupervised oceanic SAR segmentation model to bring additional information over the original vignette.

\subsection{Multi-modal Methods for SAR Semantic Segmentation}
\par The existing multi-modal methods for SAR semantic segmentation use multi-modal data (e.g., pairs of EO and SAR images) together for both training and testing\cite{CMC}, \cite{multirelate1}, \cite{RSDNet}, \cite{MCANet}, \cite{CGNet} and \cite{multirelate2}, considering that multi-modal data generally has more abundant information than single-modal data. Sun et al.\cite{CGNet} employed building footprints to learn multi-level visual features and normalize the features for predicting building masks in SAR images. Li et al. \cite{MCANet} designed a multi-modal cross attention network (MCANet) to extract multi-scale attention maps by fusing SAR and EO images. Cha et al. \cite{CMC} formulated multi-modal representation learning in contrastive multi-view coding by considering three modalities (i.e., EO image, SAR image, and label mask) as different data augmentation techniques. Jain et al. \cite{RSDNet} proposed a self-supervised method to learn invariant feature embeddings between SAR images and multi-spectral images.

\par It is worth noting that it is not a hard job to collect pairs of SAR and EO images for training these multi-modal segmentation methods offline, however, when these multi-modal methods are used for testing in many real cases (e.g., night, cloud cover, etc.) where no clear EO images but only SAR images could be captured, their performances would become significantly worse as illustrated in Fig. \ref{Figure 1} and discussed in Section \ref{1}. Unlike these multi-modal segmentation methods in literature, the proposed HFD-Net in this work focuses on a novel segmentation configuration, where pairs of SAR and EO images are used for network training, but only SAR images without EO images are used for testing. 

\section{Methodology}
\label{3}
\par In this section, we propose the Heterogeneous Feature Distillation Network (HFD-Net) for SAR image semantic segmentation, where the heterogeneous feature distillation model is explored for heterogeneous feature transfer and the heterogeneous feature alignment module is explored for multi-scale feature aggregation. Firstly, the architecture of the HFD-Net is described. Then, we present the heterogeneous feature distillation model and the heterogeneous feature alignment module respectively in detail. Finally, the model training and total loss function are introduced.

\begin{figure*}[!t]
\centerline{\includegraphics[scale=0.75]{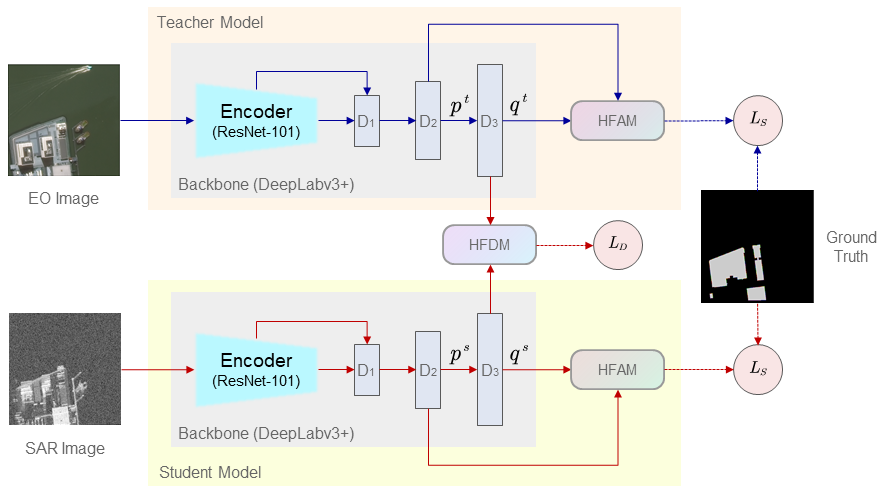}}
\caption{Architecture of HFD-Net. The HFD-Net consists of a teacher model represented in pink, a student model represented in yellow, and a heterogeneous feature distillation model (HFDM). The teacher model and the student model have an identical architecture, consisting of a backbone segmentation network (i.e., DeepLabv3+\cite{EOintroDeepLabv3+}) represented in gray and a heterogeneous feature alignment module (HFAM). The backbone DeepLabv3+\cite{EOintroDeepLabv3+} contains a ResNet-101\cite{ResNet} encoder and a decoder with three convolutional blocks (denoted as ${\rm D_1}$, ${\rm D_2}$, ${\rm D_3}$ here). $p^t$ and $p^s$ are the outputs of the ${\rm D_2}$ block in the teacher and student models respectively while $q^t$ and $q^s$ are the outputs of the ${\rm D_3}$ block in the teacher and student models respectively. The blue path indicates the process for pre-training the teacher model by minimizing the basic segmentation loss $L_S$ (defined in Eq.(6)), while the red path indicates the process for training the student model by jointly minimizing the basic segmentation loss $L_S$ and the heterogeneous distillation loss $L_D$ (defined in Eq.(2)).}
\label{Figure 2}
\end{figure*}

\subsection{Architecture}
\par The HFD-Net, whose architecture is shown in Fig. \ref{Figure 2}, consists of a pre-trained teacher model for segmenting EO images, a student model for segmenting SAR images, and a designed heterogeneous feature distillation model (HFDM) for transferring latent EO features from the teacher model to the student model. The teacher model takes EO images as its inputs, while the student model takes SAR images as its inputs. The two models have an identical architecture, which is composed of a backbone segmentation network and a designed heterogeneous feature alignment module (HFAM) for multi-scale feature aggregation, but different parameter configurations. Here, we simply use the DeepLabv3+\cite{EOintroDeepLabv3+} as the backbone segmentation network, which consists of a typical ResNet-101\cite{ResNet} encoder and a three-block decoder.

\par At the training stage, the teacher model, whose inputs are EO scene images, is firstly trained by implementing an EO image segmentation task, and it is expected to extract EO features that reserve some semantic information of the input scene images. And the parameters of this teacher model would be fixed after it has been trained. Then, the student model with SAR images as inputs is trained by implementing a SAR image segmentation task. In this training process of the student model, both the ground truth segmentation maps and the learned EO features from the teacher model are utilized jointly as the supervision information to guide the student model to learn heterogeneous features, by simultaneously minimizing the basic segmentation loss $L_{S}$ and the designed heterogeneous distillation loss $L_{D}$.

\par At the testing stage, only the student model is used for segmenting an arbitrary input SAR image without its corresponding EO image. In the following subsections, the HFDM and HFAM would be introduced respectively.

\subsection{Heterogeneous Feature Distillation Model}
\par
\begin{figure}[!t]
\centerline{\includegraphics[scale=0.67]{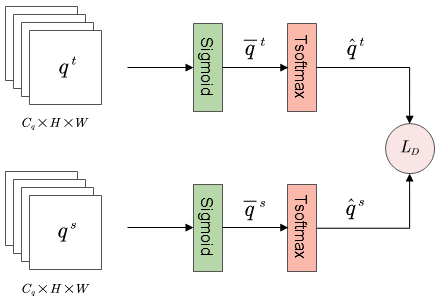}}
\caption{Architecture of HFDM. The HFDM consists of two identical Sigmoid operators and two identical Tsoftmax operators to process the input set of EO feature maps $q^t$ and the input set of SAR feature maps $q^s$ respectively, where each set of feature maps has $C_q$ channels and the size of each feature map is $\{H,W\}$. $\bar{q}^t$ and $\bar{q}^s$ denote two sets of normalized feature maps generated by Sigmoid operators. $\hat{q}^t$ and $\hat{q}^s$ denote two sets of pixel-level probability maps generated by Tsoftmax operators. $L_D$ denotes the heterogeneous distillation loss.}
\label{Figure 3}
\end{figure}

 The heterogeneous feature distillation model (HFDM) is designed to transfer latent EO features which reserve semantic information from the teacher model to the student model for segmenting SAR images. It is noted that both the teacher and student models in the existing knowledge distillation techniques\cite{hintondistill} and \cite{featuredistill} generally deal with an identical task with homogeneous images. Unlike these works, the teacher and student models in the HFD-Net focus on two similar but different tasks (one is EO image segmentation, while the other one is SAR image segmentation), hence, we design a special architecture with a heterogeneous distillation loss term for the HFDM so that the EO knowledge from the teacher model could be distilled to the SAR-segmentation student model, as shown in Fig. \ref{Figure 3}.

\par As seen from Fig. \ref{Figure 3}, the HFDM consists of two Sigmoid operators, two Tsoftmax operators and a designed heterogeneous distillation loss $L_{D}$. Given an input pair of EO and SAR images, we denote the extracted set of EO feature maps from the ${\rm D_3}$ block in the teacher model as $q^t=\{q_c^t | q_c^t \in R^{ H\times W}, c=1,2,...,C_q\}$ where $C_q$ is the number of channels in the third decoder block ${\rm D_3}$ and $\{H, W\}$ is the size of each feature map $q_c^t$. Similarly, we also denote the extracted set of SAR feature maps from the ${\rm D_3}$ block in the student model as $q^s=\{q_c^s | q_c^s \in R^{ H\times W}, c=1,2,...,C_q\}$. The HFDM is used to enforce the student model to output such SAR features $q^s$ that are as similar to the extracted EO features $q^t$ from the teacher model as possible.

\par Firstly, two identical Sigmoid operators are implemented to normalize the elements in each feature map $q_c$ (indicating both $q_c^t$ and $q_c^s$) into $(0,1)$ respectively. Then, each normalized feature map $\bar{q}_c$ (indicating both $\bar{q}_c^t$ in $\bar{q}^t$ and $\bar{q}_c^s$ in $\bar{q}^s$)
is transformed into a pixel-level probability map $\hat{q}_c$ (indicating both $\hat{q}_c^t$ in $\hat{q}^t$ and $\hat{q}_c^s$ in $\hat{q}^s$) respectively by implementing the following Tsoftmax operator:

\begin{equation}
\hat{q}_c={\rm Tsoftmax}(\bar{q}_c;T)=\frac{{\rm exp}(\frac{\bar{q}_c}{T})}{\sum \limits_{j=1}^{C_q} {\rm exp}(\frac{\bar{q}_j}{T})}
\end{equation}

\noindent where $T$ is a preseted temperature constant (here, the Tsoftmax function degrades to the commonly used Softmax function when $T$ is set to 1, as illustrated in \cite{hintondistill}). Finally, the heterogeneous distillation loss $L_D$, which measures the difference between the obtained pixel-level probability maps, is designed to distill latent features from the EO-segmentation teacher model to the SAR-segmentation student model. The heterogeneous distillation loss $L_D$ has a cross-entropy loss form, which is defined as:

\begin{equation}
L_{D}=\frac{1}{C_q}\sum\limits_{c=1}^{C_q}\sum\limits_{i,j=1}^{H,W}\hat{q}_c^t(i,j){\rm log}(\hat{q}_c^s(i,j))
\end{equation}

\noindent where $\hat{q}_c^t(i,j)$ (or $\hat{q}_c^s(i,j)$) is the element at the position $(i,j)$ in the probability map $\hat{q}_c^t$ (or $\hat{q}_c^s$), $C_q$ is the channel number and $\{H,W\}$ is the size of the input probability map.

\subsection{Heterogeneous Feature Alignment Module}
\par
\begin{figure}[!t]
\centerline{\includegraphics[scale=0.45]{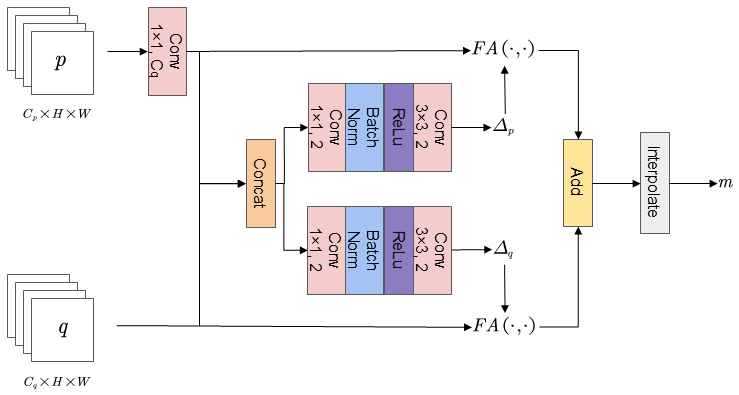}}
\caption{Architecture of HFAM. The inputs $p$ and $q$ are sets of feature maps from the ${\rm D_2}$ and ${\rm D_3}$ decoder blocks, which have $C_p$ channels and $C_q$ channels respectively, and each feature map in both $p$ and $q$ has the size of $\{H,W\}$. When the HFAM is used in the teacher model, $p$ and $q$ indicate two sets of EO feature maps $p^t$ and $q^t$, $\varDelta_p$ and $\varDelta_q$ indicate two sets of EO offset maps $\varDelta_p^t$ and $\varDelta_q^t$, $m$ indicates the predicted EO segmentation map $m^t$. When the HFAM is used in the student model, $p$ and $q$ indicate two sets of SAR feature maps $p^s$ and $q^s$, $\varDelta_p$ and $\varDelta_q$ indicate the two sets of SAR offset maps $\varDelta_p^s$ and $\varDelta_q^s$, $m$ indicates the predicted SAR segmentation map $m^s$. `${\rm FA}(\cdot,\cdot)$' is the element-level alignment function. `Conv $k\times k, N$' denotes a $k \times k$ convolutional layer with $N$ channels. `Concat' denotes channel concatenation. `BatchNorm' follows \cite{BatchNorm} paradigm. `ReLU' denotes the activation function in \cite{ReLU}. `Add' denotes pixel-level addition. `Interpolate' denotes bilinear interpolation.}
\label{Figure 4}
\end{figure}

 The heterogeneous feature alignment module (HFAM) aims to aggregate multi-scale features in both the teacher and student models. The architecture of the designed HFAM is shown in Fig. \ref{Figure 4}. 

\par As seen from Fig. \ref{Figure 4}, unlike the existing feature alignment strategies which could only aggregate homogeneous features, the HFAM employs a dual-stream architecture for aggregating heterogeneous (also homogeneous) features, where one stream is used to handle a kind of features $p \in R^{C_p \times H\times W}$ (indicating $p^t$ or $p^s$) from the second decoder block ${\rm D_2}$ of the backbone segmentation network and the other stream is used to handle another input kind of features $q\in R^{C_q \times H\times W}$ (indicating $q^t$ or $q^s$) from the third decoder block ${\rm D_3}$. Obviously, when used in the teacher model, the HFAM could aggregate homogeneous EO features. When used in the student model, the HFAM could aggregate heterogeneous EO/SAR features. Here, considering that the implementation process of HFAM in the student model is the same as that in the teacher model regardless of whether the input features are homogeneous or heterogeneous, we only introduce the implementation process of HFAM in the student model as follows. 

\par Firstly, the channel numbers of the two input features $p^s$ and $q^s$ are adjusted to be consistent by employing a 1$\times$1 convolution with $C_q$ channels to the input features $p^s$. Then, the HFAM learns two sets of offset maps $\varDelta _p^s \in R^{2\times H\times W}$ and $\varDelta _q^s \in R^{2\times H\times W}$ by concatenating the adjusted $p^s$ and $q^s$ and then passing through the dual-stream architecture respectively, where the two streams have an identical structure -- a two-channel convolution layer with 1$\times$1 kernels, a BatchNorm layer\cite{BatchNorm}, a ReLU layer\cite{ReLU} and a two-channel convolution layer with 3$\times$3 kernels -- but different parameter configurations. Next, each element in the feature map $q_c^s$ (or the adjusted feature map $p_c^s$) is aligned by the element-level feature alignment function ${\rm FA}(\cdot,\cdot)$ with the learned offset maps. After traversing all of the element in each feature map $q_c^s$ in $q^s$ (or in each adjusted feature map $p_c^s$ in adjusted $p^s$), the set of aligned feature maps $\Tilde{q}^s=\{\Tilde{q}_c^s\}_{c=1}^{C_q}$ (or $\Tilde{p}^s=\{\Tilde{p}_c^s\}_{c=1}^{C_q}$) is generated. The element-level feature alignment function ${\rm FA}(\cdot,\cdot)$ is formulated as:

\begin{equation}
    \begin{aligned}
    \Tilde{q}_c^s(i,j) &= {\rm FA}(q_c^s,\varDelta_q^s;i,j)\\
                   &= \sum\limits_{h=1}^H\sum\limits_{w=1}^W q_c^s(h,w)\\
                   &\times \max(0,1-|i+\varDelta_{q,1}^s(i,j)-h|)\\
                   &\times \max(0,1-|j+\varDelta_{q,2}^s(i,j)-w|)
    \end{aligned}
\end{equation}

\noindent where $\Tilde{q}_c^s(i,j)$ is the element at the position $(i,j)$ in the aligned feature map $\Tilde{q}_c^s$, $\varDelta_{q,1}^s(h,w)$ is the element at the position $(h,w)$ in the first offset map $\varDelta_{q,1}^s$ in $\varDelta_q^s$ and $\varDelta_{q,2}^s(h,w)$ is the element at the position $(h,w)$ in the second offset map $\varDelta_{q,2}^s$ in $\varDelta_q^s$. Eq.(3) is computed against each $p_c^s$ similarly. Finally, the two sets of aligned feature maps $\Tilde{p}^s$ and $\Tilde{q}^s$ are added by channel and the summation is interpolated under bilinear strategy to generate a predicted segmentation map $m^s$.

\subsection{Model Training and Total Loss Function}
\par The training process of the proposed HFD-Net is divided into the following two sequential stages:
\par At the first stage, the teacher model for EO image segmentation is pre-trained. As done in\cite{RSDNet}, the basic segmentation loss $L_{S}$, which measures the difference between the set of predicted segmentation maps and the corresponding ground truth segmentation maps, is straightforwardly used in this work. It contains two terms: a cross-entropy loss $L_C$ and a focal loss $L_F$\cite{Focalloss}.

\par The cross-entropy loss $L_C$ is defined as:

\begin{equation}
L_C=\frac{1}{N}\sum\limits_{i=1}^N(-y_i{\rm log}(m_i))
\end{equation}

\noindent where $N$ is the total number of the training images, $\{m_i\}_{i=1}^N$ represent the set of predicted EO segmentation maps and $\{y_i\}_{i=1}^N$ represent the set of corresponding ground truth segmentation maps.

\par The focal loss $L_F$ is defined as:

\begin{equation}
L_F=\frac{1}{N}\sum\limits_{i=1}^N(-\alpha (1-m_ie^{-y_i})^\gamma y_i{\rm log}(m_i))
\end{equation}

\noindent where $\alpha$ is a tuning parameter for controlling class balance and $\gamma$ is a tuning parameter for controlling the weight of difficult and simple samples. 

\par Then, the basic segmentation loss $L_S$ for training the teacher model is the sum of the cross-entropy loss $L_C$ and the focal loss $L_F$, as used in \cite{RSDNet}:

\begin{equation}
L_{S}=L_{C}+L_{F}
\end{equation}

\par At the second stage, the parameters of the teacher model are fixed and only the student model is trained. The loss function $L$ of the student model at this stage contains two terms: the basic segmentation loss $L_S$ in Eq.(6) and the heterogeneous distillation loss $L_D$ in Eq.(2). It is formulated as the weighted sum of $L_S$ and $L_D$:

\begin{equation}
L=L_S + \lambda L_D
\end{equation}

\noindent where $\lambda$ is a weight parameter according to $L_D$.

\section{Experiments}
\label{4}
\subsection{Datasets and Metrics}
\par In this work, we evaluate the proposed network on the following two public datasets including the SpaceNet6 dataset \cite{SpaceNet6} and the SEN12MS dataset \cite{SEN12MS}:

\par \textbf{SpaceNet6 dataset\cite{SpaceNet6}:} It contains multi-modal data including 3401 pairs of EO and SAR images, whose building footprints are provided by the public 3D Basisregistrie Adressen en Gebouwen (3DBAG) dataset\cite{3DBAG}. According to the building footprints, the building segmentation maps could be straightforwardly obtained. This dataset is used for evaluating the proposed network in a binary segmentation task (buildings and non-building regions). 2401 pairs of EO and SAR images are randomly selected from this dataset for network training, while the rest 1000 SAR images are used for network testing.

\par \textbf{SEN12MS dataset\cite{SEN12MS}:} It contains triplets of $\{$SAR image, multi-spectral image, land cover map$\}$, which are captured by Sentinel-1, Sentinel-2, and MODIS respectively. According to the multi-spectral images, the EO images could be obtained easily. As done in \cite{RSDNet}, the coarse land cover maps are replaced by the high-resolution ground truth segmentation maps released by the Data Fusion Contest (DFC) 2020\cite{DFC2020}. We use this dataset to evaluate the proposed network in a five-class segmentation task(i.e., forest, cropland, urban/built-up areas, barren terrain and water). 5128 pairs of EO and SAR images in the training set in this dataset are used for network training, and 986 SAR images in the validation set in this dataset are used for networking testing.

\par For evaluating the proposed method on the two datasets, we adopt pixel accuracy $Acc$, the mean of intersection of union $mIoU$, and the $F1$ score as the metrics, as done in \cite{CMC} and \cite{RSDNet}. We denote $P_{ij}$ as the number of pixels that belong to class $i$ but are assigned to class $j$, then the pixel accuracy $Acc$, the mean of intersection of union $mIoU$, and the $F1$ score can be formulated as:

\begin{equation}
Acc=\frac{\sum\limits_{i=0}^CP_{ii}}{\sum\limits_{i=0}^C\sum\limits_{j=0}^CP_{ij}}
\end{equation}

\begin{equation}
mIoU=\frac{1}{C}\sum\limits_{i=0}^{C}\frac{P_{ii}}{\sum\limits_{j=0}^{C}P_{ij}+\sum\limits_{j=0}^{C}P_{ji}-P_{ii}}
\end{equation}

\begin{equation}
F1=\frac{1}{C}\sum\limits_{i=0}^{C}\frac{2P_{ii}}{\sum\limits_{j=0}^{C}P_{ji}+\sum\limits_{j=0}^{C}P_{ij}}
\end{equation}

\noindent where $C+1$ is the number of classes (one of the classes is denoted as background). 

\subsection{Implementation Details}
\par We implement the HFD-Net with PyTorch\cite{PyTorch}. For the SpaceNet6 dataset\cite{SpaceNet6}, we crop and resize the input EO and SAR images into 512$\times$512, and for the SEN12MS dataset\cite{SEN12MS}, we crop and bicubicly interpolate the input EO and SAR images into 256$\times$256. At the training stage, the teacher model is firstly trained for 50 epochs by implementing an EO semantic segmentation task. Then, the teacher model is fixed and the student model is trained for 150 epochs under the supervision of both the ground truth segmentation maps and the transferred EO features. The temperature $T$ in the Tsoftmax operator in the HFDM is set to 5. As done in \cite{hintondistill}, the weight parameter $\lambda$ in the heterogeneous distillation loss $L_D$ in Eq.(7) is set to $\lambda=T^2=25$. We set $\alpha=0.5$ and $\gamma=2$ in the focal loss $L_F$ in Eq.(5). The Adam optimizer \cite{Adam} with $\beta_1=0.5$ and $\beta_2=0.999$ is used for training the two models. The batch size is set to 8, the initial learning rate is set to $10^{-4}$ and is downgraded under Poly scheduler. At the testing stage, only the student model is used for network evaluation.

\subsection{Comparative Evaluation}

\begin{table}[!t]
\centering
\caption{Quantitative comparison on the SpaceNet6 dataset\cite{SpaceNet6}. $\uparrow$ denotes that higher is better. The best results are in bold in each metric.}
\label{Tab.1} 
\begin{tabular}{C{2.5cm}|C{1.6cm}C{1.6cm}C{1.6cm}} 
\toprule 
Methods & Acc $\uparrow$ & mIoU $\uparrow$ & F1 $\uparrow$\\
\midrule 
DeepLabv3+\cite{finetuneDeepLabv3+} & 0.9794 & 0.7731 & 0.8721 \\
SegNet\cite{finetuneSegNet} & 0.9712 & 0.6872 & 0.8146 \\
HR-SAR-Net\cite{HRSARNet} & 0.9241 & 0.1724 & 0.2946 \\
Attention U-Net\cite{AttentionUNet} & 0.9194 & 0.4249 & 0.5964\\
MCANet\cite{MCANet} & 0.4489 & 0.0665 & 0.1248 \\
CMC\cite{CMC} & 0.9756 & 0.7054 & 0.8445\\
RSDNet\cite{RSDNet} & 0.9787 & 0.7652 & 0.8670 \\
\midrule
HFD-Net(ours) & \textbf{0.9845} & \textbf{0.8272} & \textbf{0.9055} \\
\bottomrule 
\end{tabular}
\end{table}

\begin{table}[!t]
\centering
\caption{Quantitative comparison on the SEN12MS data\cite{SEN12MS} where the land cover maps are replaced by the high-resolution ground truth segmentation maps provided by the DFC 2020\cite{DFC2020}.}
\label{Tab.2} 
\begin{tabular}{C{2.5cm}|C{1.6cm}C{1.6cm}C{1.6cm}} 
\toprule 
Methods & Acc $\uparrow$ & mIoU $\uparrow$ & F1 $\uparrow$\\
\midrule 
DeepLabv3+\cite{finetuneDeepLabv3+} & 0.2744 & 0.6258 & 0.2708 \\
SegNet\cite{finetuneSegNet} & 0.1837 & 0.3990 & 0.1802  \\
HR-SAR-Net\cite{HRSARNet} & 0.1431 & 0.3200 & 0.1408  \\
Attention U-Net\cite{AttentionUNet} & 0.2491 & 0.5586 & 0.2376  \\
MCANet\cite{MCANet} & 0.0621 & 0.1089 & 0.0633  \\
CMC\cite{CMC} & 0.2693 & 0.6154 & 0.2694 \\
RSDNet\cite{RSDNet} & 0.2543 & 0.6266 & 0.2692 \\
\midrule
HFD-Net(ours) & \textbf{0.2789} & \textbf{0.6474} & \textbf{0.2775} \\
\bottomrule 
\end{tabular}
\end{table}

\par Here, seven state-of-the-art SAR image semantic segmentation methods are also evaluated for comparison, including four single-modal SAR segmentation methods (DeepLabv3+\cite{finetuneDeepLabv3+}, SegNet\cite{finetuneSegNet}, HR-SAR-Net\cite{HRSARNet}, Attention U-Net \cite{AttentionUNet}) and three multi-modal SAR segmentation methods (MCANet\cite{MCANet}, CMC\cite{CMC}, RSDNet\cite{RSDNet}). It is noted that the DeepLabv3+\cite{finetuneDeepLabv3+} also serves as the baseline model in this paper. Since no codes for these comparative methods are released currently and all these methods are not originally designed for handling the segmentation configuration (i.e., pairs of EO and SAR images for training, while only SAR images for testing) as done in this work, we evaluate these methods with our reproduced codes. The corresponding results by all the comparative methods on the two datasets are reported in Tables \ref{Tab.1}-\ref{Tab.2} respectively. It has to be pointed out that since only SAR images are available at the testing stage in our task, all the multi-modal methods (MCANet\cite{MCANet}, CMC\cite{CMC} and RSDNet\cite{RSDNet} which use additional data at their testing stage in their original papers) are also tested by utilizing SAR images singly for making a fair comparison here. This is the reason of why our reported results on these methods are different from those in their original papers.

\par As seen from Table \ref{Tab.1} where the comparative results on the SpaceNet6 dataset\cite{SpaceNet6} for handling a binary segmentation task are reported. Deeplabv3+\cite{finetuneDeepLabv3+} that uses SAR images singly for training and testing performs better than the three multi-modal methods\cite{CMC}, \cite{RSDNet} and \cite{MCANet}, mainly due to the fact that these multi-modal methods are originally designed to use multi-modal information for testing and their testing performances have to be degraded to some extent when only SAR images are accessible. In addition, the multi-modal method MCANet\cite{MCANet} achieves a lower performance than the other two multi-modal methods CMC\cite{CMC} (EO images and ground truth segmentation maps are used as multi-modal data) and RSDNet\cite{RSDNet} (multi-spectral images are used as multi-modal data), mainly because unlike CMC\cite{CMC} and RSDNet\cite{RSDNet} where the multi-modal information is used for data augmentation and loss calculation, MCANet\cite{MCANet} straightforwardly fuses pairs of EO and SAR images at the beginning, meaning that although the performances of the three methods are all dependent on extra data (i.e., EO images, multi-spectral images, ground truth segmentation maps), the dependency of MCANet\cite{MCANet} on extra data (i.e., EO images) is even stronger. Moreover, the $Acc$ of the proposed HFD-Net is 0.9845\%, slightly higher than those of DeepLabv3+\cite{finetuneDeepLabv3+}, SegNet\cite{finetuneSegNet}, CMC\cite{CMC}, and RSDNet\cite{RSDNet}, because most of the wrongly segmented pixels by these methods locate around the building contours and accordingly the number of the wrongly segmented pixels is much smaller than the total number of pixels in an input image. The proposed HFD-Net significantly outperforms all the comparative methods under the metrics $mIoU$ and $F1$, mainly due to the designed HFDM and HFAM.

\begin{figure*}[!t]
\centerline{\includegraphics[scale=0.75]{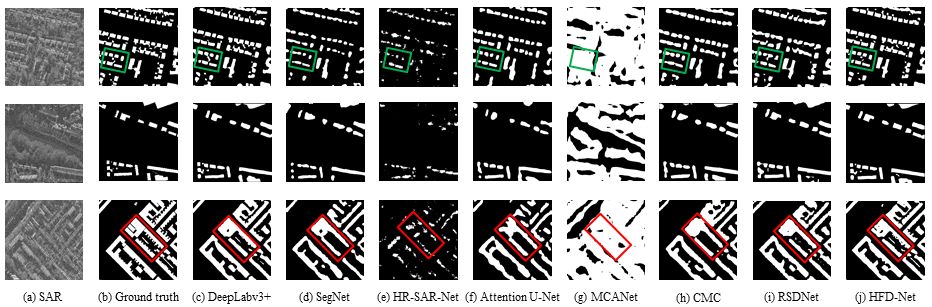}}
\caption{Visualization results on the SpaceNet6 dataset\cite{SpaceNet6}: (a) SAR images; (b) Ground truth segmentation maps; (c) Predictions by our baseline model DeepLabv3+\cite{finetuneDeepLabv3+}; (d) Predictions by SegNet\cite{finetuneSegNet}; (e) Predictions by HR-SAR-Net\cite{HRSARNet}; (f) Predictions by Attention U-Net\cite{AttentionUNet}; (g) Predictions by MCANet\cite{MCANet}; (h) Predictions by CMC\cite{CMC}; (i) Predictions by RSDNet\cite{RSDNet}; (j) Predictions by our HFD-Net.}
\label{Figure 5}
\end{figure*}

\begin{figure*}[!t]
\centerline{\includegraphics[scale=0.75]{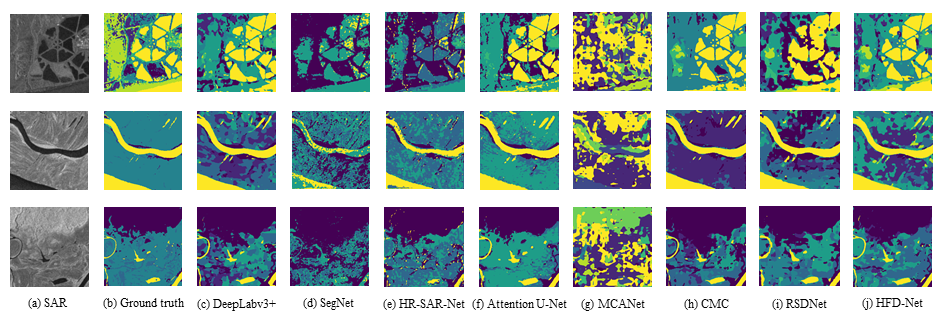}}
\caption{Visualization results on the SEN12MS dataset\cite{SEN12MS}: (a) SAR images; (b) Ground truth segmentation maps; (c) Predictions by our baseline model DeepLabv3+\cite{finetuneDeepLabv3+}; (d) Predictions by SegNet\cite{finetuneSegNet}; (e) Predictions by HR-SAR-Net\cite{HRSARNet}; (f) Predictions by Attention U-Net\cite{AttentionUNet}; (g) Predictions by MCANet\cite{MCANet}; (h) Predictions by CMC\cite{CMC}; (i) Predictions by RSDNet\cite{RSDNet}; (j) Predictions by our HFD-Net.}
\label{Figure 6}
\end{figure*}

\begin{table}[!t]
\centering
\caption{Quantitative comparison of the HFD-Net with different architecture components on the SpaceNet6 dataset\cite{SpaceNet6}.} 
\label{Tab.3} 
\begin{tabular}{C{1.6cm}C{1.6cm}C{1.6cm}|C{1.6cm}} 
\toprule 
Backbone & HFDM & HFAM & mIoU $\uparrow$\\
\midrule 
\Checkmark & & & 0.7731 \\
\Checkmark & \Checkmark & & 0.8140\\
\Checkmark & & \Checkmark & 0.8141\\
\Checkmark & \Checkmark & \Checkmark & \textbf{0.8272} \\
\bottomrule 
\end{tabular}
\end{table}

\begin{table}[!t]
\centering
\caption{Quantitative comparison of the HDF-Net with different configurations for the HFDM and HFAM on the SpaceNet6 dataset\cite{SpaceNet6}. `${\rm D_a}$+${\rm D_b}$' denotes the configuration that distills features from Db and aggregates features generated by ${\rm D_a}$ and ${\rm D_b}$.}
\label{Tab.4} 
\begin{tabular}{m{2.4cm}<{\centering}|m{1.6cm}<{\centering}m{1.6cm}<{\centering}m{1.6cm}<{\centering}} 
\toprule 
Settings & Acc $\uparrow$ & mIoU $\uparrow$ & F1 $\uparrow$\\
\midrule 
${\rm D_1}$+${\rm D_2}$ & 0.9825 & 0.8037 & 0.8912 \\
${\rm D_1}$+${\rm D_3}$ & 0.9824 & 0.8027 & 0.8905 \\
${\rm D_1}$+${\rm D_2}$+${\rm D_3}$ & 0.9834 & 0.8143 & 0.8977 \\
 HFD-Net (${\rm D_2}$+${\rm D_3}$) & \textbf{0.9845} & \textbf{0.8272} & \textbf{0.9055} \\
\bottomrule 
\end{tabular}
\end{table}

\begin{table}[!t]
\centering
\caption{Quantitative comparison of the HFD-Net with different values of the temperature $T$ on the SpaceNet6 dataset\cite{SpaceNet6}.} 
\label{Tab.5} 
\begin{tabular}{C{1.6cm}|C{1.6cm}C{1.6cm}C{1.6cm}} 
\toprule 
T & Acc $\uparrow$ & mIoU $\uparrow$ & F1 $\uparrow$\\
\midrule 
1 & 0.9836 & 0.8165 & 0.8990 \\
5 & \textbf{0.9845} & \textbf{0.8272} & \textbf{0.9055} \\
10 & 0.9834 & 0.8142 & 0.8976 \\
50 & 0.9831 & 0.8113 &  0.8958 \\
\bottomrule 
\end{tabular}
\end{table}

\par In addition, several visualization samples by all the comparative methods trained on the SpaceNet6 dataset\cite{SpaceNet6} are shown in Fig. \ref{Figure 5}. As seen from this figure, the SAR segmentation maps predicted by our HFD-Net depict more precise building contours similar to the corresponding ground truth segmentation maps, in comparison to the seven comparative methods. For example, compared with these comparative methods, small buildings (e.g., the green rectangle region in the Fig. \ref{Figure 5}) are segmented up independently by our HFD-Net rather than be segmented as a whole building by the comparative methods in most cases, and fine edges of irregular buildings (e.g., the red rectangle region in Fig. \ref{Figure 5}) are segmented more accurately according to the corresponding ground truth segmentation maps by the proposed HFD-Net.

\par Furthermore, Table \ref{Tab.2} reports the comparative results on the SEN12MS dataset\cite{SEN12MS} for handling a multi-class segmentation task. It is noted from Table \ref{Tab.2} that the HFD-Net also outperforms all the other comparative methods, which is consistent with the results on the SpaceNet6 dataset\cite{SpaceNet6} reported in Table \ref{Tab.1}. These results further demonstrate the effectiveness of the proposed HFD-Net. Fig. \ref{Figure 6}  shows several visulization results by all the comparative methods trained on the SEN12MS dataset\cite{SEN12MS}. As seen from this figure, the HFD-Net could segment SAR images into various classes more effectively than the other methods.

\subsection{Ablation Studies}
\par This subsection verifies the effectiveness of each key element in the HFD-Net by conducting ablation studies on the two datasets. 

\par In order to demonstrate the effectiveness of the HFDM and HFAM, we sequentially trained the following three models on the SpaceNet6 dataset\cite{SpaceNet6}: (i) the baseline model which only contains the backbone DeepLabv3+\cite{EOintroDeepLabv3+} without the HFDM and HFAM, (ii) the model consisting of the baseline model and the HFDM, and (iii) the model consisting of the baseline model and the HFAM, and compare these models with the proposed model (consisting of the baseline model, HFDM and HFAM). The corresponding results are reported in Table \ref{Tab.3}. These results show that both the HFDM and HFAM are helpful for improving the segmentation performance. And the performance can be further improved when using the HFDM and HFAM together (which is done in the proposed HFD-Net).

\par In order to further verify different possible configurations for the HFDM and HFAM, we trained several models on the SpaceNet6 dataset\cite{SpaceNet6} where the HFDM distills features and the HFAM aggregates features under the following four configurations: (i) distilling features from the decoder block ${\rm D_2}$ in the teacher model to the student model and aggregating features generated by the decoder blocks ${\rm D_1}$ and ${\rm D_2}$ in the student/teacher models, denoted as ``${\rm D_1}$+${\rm D_2}$"; (ii) distilling features from the decoder block ${\rm D_3}$ and aggregating features generated by the decoder blocks ${\rm D_1}$ and ${\rm D_3}$, denoted as ``${\rm D_1}$+${\rm D_3}$"; (iii) distilling features from the decoder blocks ${\rm D_1}$, ${\rm D_3}$ and aggregating features generated by the decoder blocks ${\rm D_1}$, ${\rm D_2}$, and ${\rm D_3}$, denoted as ``${\rm D_1}$+${\rm D_2}$+${\rm D_3}$"; (iv) distilling features from the decoder block ${\rm D_3}$ and aggregating features generated by the decoder blocks ${\rm D_2}$ and ${\rm D_3}$, denoted as ``${\rm D_2}$+${\rm D_3}$" (which is the configuration used in the proposed HFD-Net). The corresponding results are reported in Table \ref{Tab.4}. As seen from this table, the configuration ``${\rm D_2}$+${\rm D_3}$" used in the proposed HFD-Net is slightly better than the other configurations. Additionally, as seen from both Table \ref{Tab.4} and Table \ref{Tab.1}, no matter which configuration for the HFDM and HFAM is used, the proposed method always performs better than the comparative methods, which could further demonstrate the effectiveness of the designed HFDM and HFAM.

\par
\begin{figure}[!t]
\centerline{\includegraphics[scale=0.55]{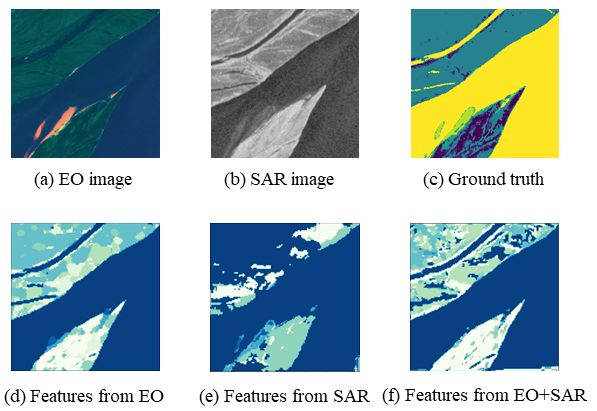}}
\caption{Effectiveness of the HFDM for transferring heterogeneous features. Three models are trained on the SEN12MS dataset\cite{SEN12MS}: (i) the teacher model; (ii) the student model without EO images; (iii) the proposed HFD-Net where EO images are used. And (a) is an EO image; (b) is the SAR image according to the EO image (a); (c) is the corresponding ground truth segmentation map; (d) is the feature map obtained by model (i); (e) is the feature map obtained by model (ii) where only SAR images are used for training; (f) is the feature map obtained by model (iii) where both EO and SAR images are used for training.}
\label{Figure 7}
\end{figure}

\begin{figure}[!t]
\centerline{\includegraphics[scale=0.5]{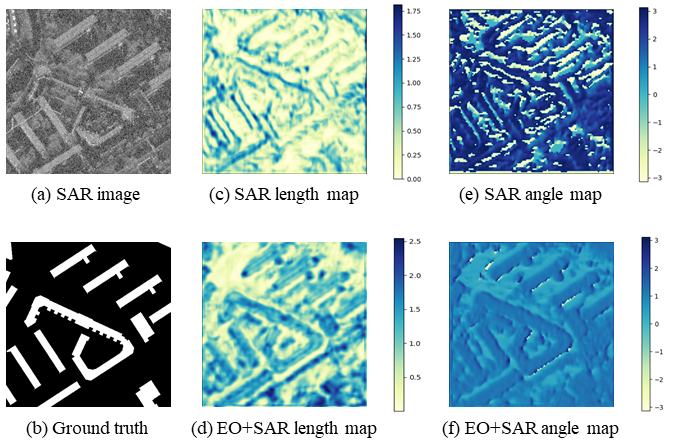}}
\caption{Effectiveness of latent EO features for heterogeneous feature alignment in the HFAM. Two models are trained on the SpaceNet6 dataset\cite{SpaceNet6}: (i) the student model without latent EO features and (ii) the proposed HFD-Net with latent EO features. And (a) is a SAR image; (b) is the ground truth segmentation map according to the SAR image (a); (c) and (e) are the length map and angle map from the set of offset maps generated by model (i); (d) and (f) are the length map and angle map from the set of offset maps generated by model (ii).}
\label{Figure 8}
\end{figure}

 In order to further illustrate the effectiveness of the latent EO features distilled from the HFDM, we visualize the feature maps and offset maps in Fig. \ref{Figure 7}-\ref{Figure 8}. Firstly, the effectiveness of the HFDM for transferring heterogeneous features is explored in Fig. \ref{Figure 7}. We trained the following three models on the SEN12MS dataset\cite{SEN12MS} respectively: (i) the teacher model alone for EO segmentation; (ii) the student model alone for SAR segmentation where the HFDM is not used; (iii) the proposed HFD-Net for SAR segmentation where the HFDM is used to transfer latent EO features from the teacher model to the student model, and visualize the feature maps from the decoder block ${\rm D_3}$ in these models (i.e., (d)-(f) in Fig. \ref{Figure 7}). As seen from Fig. \ref{Figure 7}, the feature map generated by the proposed HFD-Net ((f) in Fig. \ref{Figure 7}) is more similar with the EO feature map generated by the independently trained teacher model ((d) in Fig. \ref{Figure 7}) than the feature map generated by the independently trained student model ((e) in Fig. \ref{Figure 7}), indicating the effectiveness of the designed HFDM for feature distillation and feature transferring. Then, the effectiveness of the latent EO features for heterogeneous feature alignment in the HFAM is further explored in Fig. \ref{Figure 8}. We trained the following two models on the SpaceNet6 dataset\cite{SpaceNet6}: (i) the student model without EO images and (ii) the proposed HFD-Net where the EO images are used for training, and obtain two sets of offset maps $\varDelta_p^s$ respectively. The length maps ((c) and (d) in Fig. \ref{Figure 8}) representing the length of the offset vector at each pixel in the feature map $p_c^s$, and the angle maps ((e) and (f) in Fig. \ref{Figure 8}) representing the angle of the offset vector at each pixel can be straightforwardly obtained from the sets of offset maps $\varDelta_p^s$. As seen from Fig. \ref{Figure 8}, the length map and angle map obtained by the proposed HFD-Net where EO images are used ((d) and (f) in Fig. \ref{Figure 8}) depict the shape of the buildings clearly by assigning larger value to pixels within the buildings and lower value to pixels at the building contours, according to the ground truth segmentation map ((b) in Fig. \ref{Figure 8}), while the length map and angle map obtained by the student model without EO images ((c) and (e) in Fig. \ref{Figure 8}) assign low length value and high angle value to all pixels in the feature map $p_c^s$. Thus, this figure illustrates the effectiveness of latent EO features for heterogeneous feature alignment in the student model.

\par Finally, we evaluate the influence of the temperature $T$ in Eq.(1) by training the HFD-Net with $T={1,5,10,50}$ respectively. The corresponding results are reported in Table \ref{Tab.5}. As seen from this table, the performance rises and then falls a little, reaching the peak at $T=5$. Since our samples are discrete, we can only describe the rough curve, and the peak value may not be exact. Since the results are close in Table \ref{Tab.5}, we simply pick 5 for $T$.

\section{Conclusions}
\label{5}
\par In this paper, we propose the HFD-Net for single SAR image semantic segmentation, which contains an EO-segmentation teacher model, a SAR-segmentation student model and a heterogeneous feature distillation model. The teacher model and the student model have an identical architecture but different parameter configurations, and the heterogeneous feature distillation model is used for feature transfer. We also explore a heterogeneous feature alignment module for multi-scale feature aggregation. Experimental results on two public datasets demonstrate the effectiveness of the proposed model in comparison to seven state-of-the-art methods for SAR semantic segmentation.

\par In future, we would further investigate how to utilize EO features to boost SAR semantic segmentation more effectively, considering that EO images could provide more abundant textures and structural information for semantic segmentation.

\bibliography{ref_tgrs}
\end{document}